\documentclass{article}
\usepackage{spconf,amsmath,graphicx,booktabs,hyperref}

\title{FlowAtom: Atom-Based Evidence Aggregation for Multi-Label Website Fingerprinting}
\name{
Chongru Fan$^{1,2}$,
Wentao Huang$^{1}$,
Wei Wang$^{2}$,
Zhenquan Ding$^{2,*}$,
Jinqiao Shi$^{1,*}$,
Wei Cai$^{2}$,
Zhiyu Hao$^{2}$
\thanks{$^{*}$Corresponding authors: Zhenquan Ding (dingzq@zgclab.edu.cn) and Jinqiao Shi (shijinqiao@bupt.edu.cn).}
\thanks{This work was supported by the National Major Science and Technology
Project for Cyberspace Security under Grant 2025ZD1501502.}
}

\address{
$^{1}$School of Cyberspace Security, Beijing University of Posts and Telecommunications, Beijing, China\\
$^{2}$Zhongguancun Laboratory, Beijing, China
}

\begin{document}
%
\maketitle
\begin{abstract}

   Identifying the set of monitored websites in mixed encrypted traffic is challenging because an individual flow often provides only partial evidence of website identity. To address this challenge, we propose FlowAtom, which constructs shared prototypes, called Atoms, from flow representations without website labels. Specifically, FlowAtom pretrains a flow encoder on external unlabeled traffic and aggregates Atom responses across flows within each observation window into a fixed-dimensional, permutation-invariant representation for monitored website-set prediction. Across Direct HTTPS, Trojan, and VMess, FlowAtom achieves micro-F1 scores of 97.82\%, 94.43\%, and 93.92\% in closed-world evaluation, respectively, and consistently outperforms the evaluated baselines in open-world evaluation on windows containing monitored visits. The code is available at \url{https://github.com/aimafan123/FlowAtom}.

\end{abstract}
\section{Introduction}
\label{sec:intro}

Website fingerprinting (WF) attacks infer the websites a user visits from side-channel features of encrypted traffic~\cite{wang2014effectiveAttacks, hayes2016kFingerprinting, li2023robust,Cheng2025STARSA}. Traditional WF studies typically use the complete traffic trace of a single website visit as one sample and formulate website identification as single-label classification~\cite{sirinam2018deepFingerprinting,bhat2019varcnn,rahman2020tiktok}. In realistic browsing, however, multi-tab activity can generate overlapping visits to multiple websites, while background services and unmonitored websites contribute additional traffic~\cite{critical2014juarez,jin2023multitabTransformer,flowatom_R10}. These multi-website scenarios motivate website-set identification: inferring the set of websites represented in mixed traffic~\cite{deng2023robustMultitab,jin2023multitabTransformer,meng2025beyondSingleTabs}.

Packet-sequence-based multi-label WF methods, such as~\cite{jin2023multitabTransformer,deng2023robustMultitab}, take predefined traffic traces as input and predict website labels from packet sequences. For approaches that require predefined visit traces, processing continuous traffic depends on packet-level trace segmentation to locate those traces; visit boundaries can be difficult to recover accurately~\cite{wang2016realistically}. In the settings considered here, individual connections are distinguishable, allowing an attacker to separate flows using observable connection identifiers~\cite{ma2024encryptedProxies}. This groups packets by connection, without determining which website visit generated each flow. We therefore take a flow-level perspective: the model receives an unordered flow set for each observation window and predicts the monitored website set without packet-level trace segmentation by visit or flow-to-website assignment at inference. Our evaluation constructs these windows offline from pre-segmented visit traces.

\begin{figure}[!t]
    \centering
    \includegraphics[width=\columnwidth]{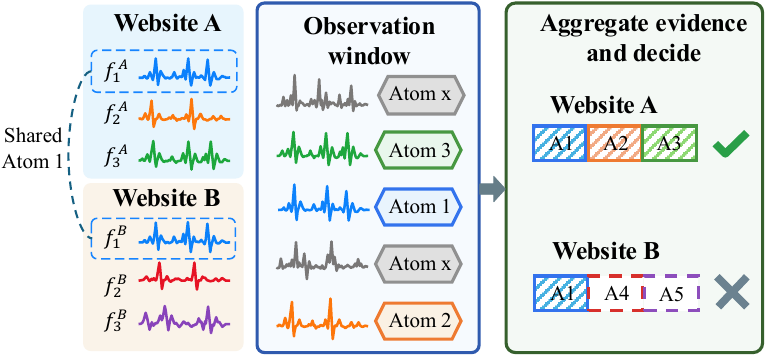}
    \caption{Different websites can generate similar flows; aggregating complementary evidence across flows in an observation window supports prediction of the monitored website set.}
    \label{fig:motivation}
\end{figure}

As illustrated in Fig.~\ref{fig:motivation}, an individual flow often provides only partial evidence of website identity~\cite{ma2024encryptedProxies,li2023robust}. A website visit typically generates multiple flows, while shared content delivery networks (CDNs), third-party services, and resources can produce similar flow characteristics across different websites~\cite{Doan2022Empirical,demystifying2013xiao,ma2024encryptedProxies}. Consequently, a flow may contain only part of the traffic generated by a website visit and may not identify the website on its own. Discriminative evidence may therefore be distributed across multiple flows within an observation window. We represent shared patterns in the flow representation space using prototypes called Atoms, and aggregate their responses to predict the monitored website set for an observation window.

\begin{figure*}[t]
    \centering
    \includegraphics[width=\textwidth]{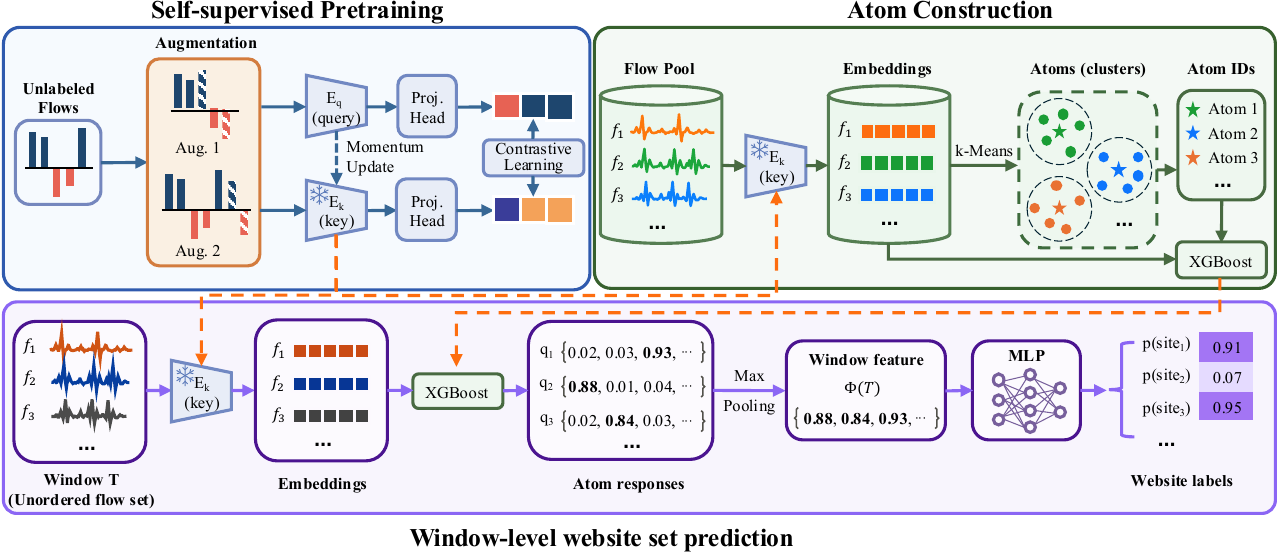}
    \caption{Overview of FlowAtom: flow representation pretraining, Atom construction, and window-level multi-label prediction.}
    \label{fig:framework}
\end{figure*}

Building on this intuition, we propose FlowAtom, which constructs shared Atoms by clustering flow representations without using website labels and represents each flow by a vector of Atom response scores. For each observation window, FlowAtom applies max pooling to each Atom's response scores across flows, producing a fixed-dimensional, permutation-invariant window representation for predicting the monitored website set~\cite{flowatom_R06}.

Our contributions are as follows. First, we formulate multi-label WF from a flow-level perspective, using an unordered flow set as input without requiring packet-level trace segmentation by visit or flow-to-website assignment at inference. We construct an evaluation benchmark from offline mixtures of pre-segmented visit traces under Direct HTTPS and non-multiplexed encrypted proxies. Second, we propose FlowAtom, which constructs shared Atoms without website-label supervision and aggregates their response scores across flows into a fixed-dimensional, permutation-invariant window representation for multi-label prediction. Third, we evaluate FlowAtom under Direct HTTPS, Trojan~\cite{trojan_gfw_2023}, and VMess~\cite{project_x_2020}, obtaining closed-world micro-F1 scores of 97.82\%, 94.43\%, and 93.92\% and open-world micro-F1 scores of 92.37\%, 92.64\%, and 89.85\%, respectively.

\section{Threat Model}
\label{sec:threat}

We consider Direct HTTPS and non-multiplexed encrypted proxy settings in which the attacker can distinguish individual connections. The attacker observes traffic on the client egress link in the Direct HTTPS setting and on the client–proxy link in the encrypted proxy settings. This passive attacker uses five-tuples or equivalent connection identifiers to separate flows and observes packet payload lengths and directions, but cannot decrypt or modify the traffic. Connection identifiers are used only for flow separation; the model input excludes IP addresses, DNS information, and Server Name Indication (SNI). For each observation window, the model receives an unordered flow set, without website-visit boundaries or flow-to-website assignments. The attacker must still separate flows by connection; the model does not require packet-level trace segmentation by visit.

\section{Method}
\label{sec:method}

As shown in Fig.~\ref{fig:framework}, FlowAtom comprises three stages: flow representation pretraining, Atom construction, and window-level multi-label prediction.

During pretraining, FlowAtom removes zero-payload packets and represents each flow as a sequence of signed packet payload lengths, preserving packet order and using the sign to encode direction. Each sequence is truncated or padded to $L=300$ packet positions. We use these sequences from large-scale external unlabeled traffic for MoCo-style contrastive pretraining~\cite{flowatom_R07}. For each input flow, we randomly select two distinct augmentation operations from ~\cite{Xie2023Rosetta} to generate two views for contrastive learning of flow representations. The encoder $E$ maps each input flow $f_i$ to a representation $\mathbf{z}_i = E(f_i)$. After pretraining, we discard the projection head and freeze the encoder $E$ for Atom construction and all subsequent window representation computations.

For each target traffic scenario, FlowAtom constructs an Atom space using only flows from that scenario's training split, without using their website labels. We first extract training-flow representations using the frozen encoder $\mathbf{z}_i = E(f_i)$. We then apply k-means clustering~\cite{flowatom_R11} to these training-flow representations and discard clusters with fewer members than a minimum cluster-size threshold. The centers of the retained clusters form a set of shared prototypes, called Atoms: $\mathcal{A}=\{a_1,a_2,\dots,a_A\}$, where $A$ is the number of retained clusters. We train an XGBoost mapper $M$~\cite{flowatom_R08} using the k-means cluster assignments as pseudo-labels. The mapper outputs an Atom response vector $\mathbf{q}_i=M(\mathbf{z}_i)$, where $q_{i,a}$ is the response score of flow $f_i$ for Atom $a$. These Atoms are learned without website labels, allowing flows from different website visits to be represented in a shared Atom space. The k-means algorithm is used only for offline Atom construction. For subsequent window representation computation and inference, both $E$ and $M$ remain frozen and are applied sequentially to compute Atom responses.

To construct training and evaluation windows offline, we select one visit trace from each of one or more websites within the same data split and combine their valid flows into an unordered flow set $T$. In this multi-instance multi-label formulation~\cite{flowatom_R05}, flows are instances, each window is a bag of flows, and the monitored websites present in the window form its label set. For each flow in $T$, we compute its Atom response vector using the frozen encoder $E$ followed by the frozen mapper $M$. We then apply max pooling over flows for each Atom~\cite{flowatom_R13}: $\Phi_a(T)=\max_{f_i\in T}q_{i,a}$. This yields an $A$-dimensional window representation that is invariant to flow order, where $A$ is the number of retained Atoms. We estimate the standardization parameters for window representations using only downstream training windows. We then train a multilayer perceptron (MLP) decoder with two hidden layers using window-level website labels and binary cross-entropy with positive-class weighting. During decoder training, $E$ and $M$ remain frozen; the validation set is used for model and decision-threshold selection. At inference time, we apply flow encoding, Atom mapping, max pooling, standardization, and multi-label decoding in sequence to the input flow set to predict the monitored websites present in the window.

\section{Evaluation}
\subsection{Evaluation Setup}

The experimental data comprise unlabeled traffic for pretraining, monitored website traffic, and unmonitored website traffic used as background traffic. For pretraining, we sample 1,000,000 unlabeled flows from JP-MAWI~\cite{flowatom_R12}. For monitored traffic, we randomly select 100 websites from the Tranco Top 10K~\cite{lepochat2019tranco} and collect website-visit traces under Direct HTTPS, Trojan, and VMess. For open-world evaluation, we exclude the 100 monitored websites from the same list and collect one visit trace for each remaining website, which we treat as unmonitored. Unmonitored website traffic is used only for testing and is excluded from model training and selection.

We compare FlowAtom with ARES~\cite{deng2023robustMultitab}, BAPM~\cite{flowatom_R10}, TMWF~\cite{jin2023multitabTransformer}, CAWF~\cite{ma2024encryptedProxies}, and Flow-DF, a flow-based adaptation of DF~\cite{sirinam2018deepFingerprinting}. ARES uses Transformer-based classifiers to identify websites from local patterns within traffic segments. BAPM combines convolution and attention mechanisms for website identification, whereas TMWF uses a Transformer. Flow-DF encodes each flow’s signed payload-length sequence to predict website scores, then aggregates these scores across flows to obtain window-level predictions. CAWF predicts flow classes from per-flow statistical features and identifies websites using contextual relationships within the resulting flow-class sequences. All methods use the same data splits, constructed observation windows, and evaluation metrics. The baseline implementations are based on the corresponding published methods.

For each traffic scenario, we split single-website visit traces into disjoint training, validation, and test sets. Within each split, we construct observation windows offline by combining the flows from complete, pre-segmented visit traces. Each window contains visits to $m\in\{1,2,3,4,5\}$ distinct monitored websites, where $m$ counts monitored websites only. FlowAtom receives only the mixed unordered flow set; website-visit boundaries, flow-to-website assignments, and the true number of monitored websites $m$ are not provided. In closed-world testing, $m=1$ denotes the single-website condition, and $m\in\{2,3,4,5\}$ denotes the multi-website condition. Within each traffic scenario, we evaluate the same trained model under both conditions. For open-world testing, we add the flows from $b\in\{1,2,3,4,5\}$ unmonitored website-visit traces to each window containing monitored visits and evaluate the resulting performance in identifying the monitored website set.

We use micro-averaged F1 (micro-F1) as the primary evaluation metric. For each evaluation task, we perform five runs with different random seeds and report the mean micro-F1 across the five runs.

\subsection{Closed-World Experiments}

\begin{table*}[t]
    \centering
    \caption{Closed-world micro-F1 (\%) in three traffic scenarios, with $m\in\{1,2,3,4,5\}$ distinct monitored websites per observation window. Bold and underlined values indicate the best and second-best results in each column, respectively.}
    \label{tab:closed_sm}
    {\small

        \setlength{\tabcolsep}{2pt}

        \begin{tabular*}{\textwidth}{@{\extracolsep{\fill}}l*{15}{c}@{}}

            \toprule

            & \multicolumn{5}{c}{Direct HTTPS}

            & \multicolumn{5}{c}{Trojan}

            & \multicolumn{5}{c}{VMess} \\

            \cmidrule(lr){2-6}\cmidrule(lr){7-11}\cmidrule(lr){12-16}

            Method

            & $1$ & $2$ & $3$ & $4$ & $5$

            & $1$ & $2$ & $3$ & $4$ & $5$

            & $1$ & $2$ & $3$ & $4$ & $5$ \\

            \midrule

            ARES
            & $67.91$ & $57.14$ & $40.49$ & $37.41$ & $35.77$ & \underline{$88.20$} & $73.37$ & $62.42$ & $57.07$ & $53.73$ & \underline{$89.89$} & $74.74$ & $63.73$ & $61.79$ & $53.82$ \\

            BAPM
            & $57.57$ & $34.21$ & $25.30$ & $21.05$ & $15.66$ & $64.17$ & $34.33$ & $33.61$ & $32.39$ & $23.73$ & $65.26$ & $37.11$ & $31.71$ & $30.95$ & $24.12$ \\

            TMWF
            & $68.85$ & $35.72$ & $29.39$ & $25.45$ & $20.22$ & $79.42$ & $42.38$ & $36.84$ & $36.55$ & $30.00$ & $82.83$ & $46.91$ & $39.67$ & $39.98$ & $32.53$ \\

            Flow-DF
            & $75.92$ & $80.20$ & $78.82$ & $78.66$ & $81.72$ & $82.29$ & $79.62$ & $77.05$ & \underline{$81.30$} & \underline{$82.07$} & $82.56$ & \underline{$82.49$} & \underline{$82.21$} & \underline{$84.23$} & \underline{$80.10$} \\

            CAWF
            & \underline{$84.72$} & \underline{$86.59$} & \underline{$86.06$} & \underline{$81.00$} & \underline{$84.19$} & $82.41$ & \underline{$84.75$} & \underline{$81.87$} & $80.80$ & $80.52$ & $80.85$ & $79.85$ & $77.39$ & $81.79$ & $73.62$ \\

            \textbf{FlowAtom}
            & $\boldsymbol{99.35}$ & $\boldsymbol{98.76}$ & $\boldsymbol{96.71}$ & $\boldsymbol{97.70}$ & $\boldsymbol{96.78}$ & $\boldsymbol{95.57}$ & $\boldsymbol{96.26}$ & $\boldsymbol{95.00}$ & $\boldsymbol{92.37}$ & $\boldsymbol{93.08}$ & $\boldsymbol{94.53}$ & $\boldsymbol{93.83}$ & $\boldsymbol{94.58}$ & $\boldsymbol{93.66}$ & $\boldsymbol{92.86}$ \\

            \bottomrule

        \end{tabular*}

    }

\end{table*}

In the closed-world setting, we evaluate multi-label WF under single-website ($m=1$) and multi-website ($m\in\{2,3,4,5\}$) conditions. Table~\ref{tab:closed_sm} shows that FlowAtom achieves the highest micro-F1 among the evaluated methods in all 15 closed-world settings, outperforming the strongest baseline in each setting by 4.64--16.70 percentage points.

\subsection{Open-World Experiments}

\begin{figure}[!t]
    \centering
    \includegraphics[width=\columnwidth]{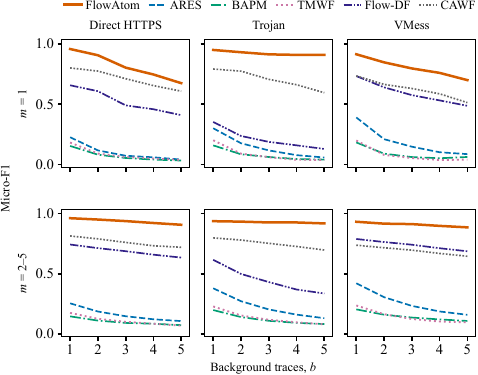}
    \caption{Open-world micro-F1 as the number of unmonitored visit traces $b$ increases. Results are reported separately for windows with one monitored website ($m=1$) and pooled windows with multiple monitored websites ($m\in\{2,3,4,5\}$).}
    \label{fig:open_world}
\end{figure}

Fig.~\ref{fig:open_world} reports results for 30 combinations of three traffic scenarios, two monitored-website conditions ($m=1$ and pooled $m\in\{2,3,4,5\}$), and five values of b. All evaluated windows contain monitored visits, and FlowAtom achieves the highest micro-F1 among the evaluated methods in every combination.

\subsection{Ablation Experiments}

We separately vary the parameter $K$ and the number of flows retained per visit trace under Direct HTTPS, Trojan, and VMess to examine how performance changes with each parameter.

\begin{figure}[!t]
    \centering
    \includegraphics[width=\columnwidth]{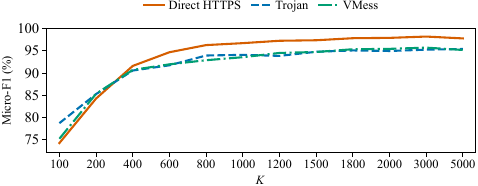}
    \caption{Effect of the parameter $K$ on micro-F1 (\%) under Direct HTTPS, Trojan, and VMess.}
    \label{fig:atom}
\end{figure}

Fig.~\ref{fig:atom} shows that micro-F1 generally increases with $K$ in all three traffic scenarios and then plateaus with small fluctuations at larger values.

\begin{figure}[!t]
    \centering
    \includegraphics[width=\columnwidth]{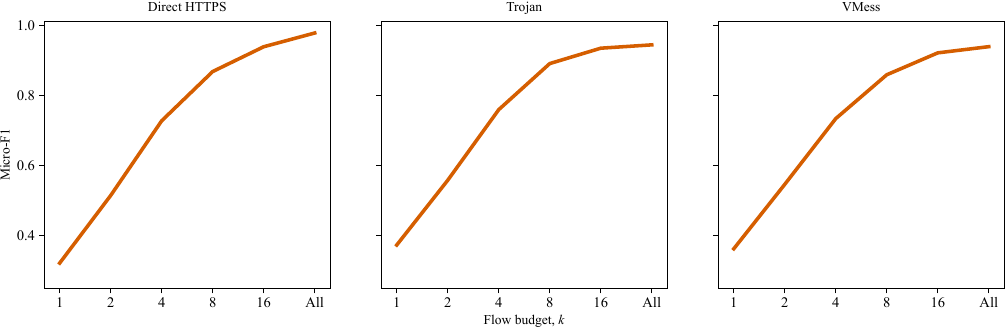}
    \caption{Effect of the number of flows retained per visit trace, $k$, on micro-F1 under Direct HTTPS, Trojan, and VMess.}
    \label{fig:flow_budget}
\end{figure}

Fig.~\ref{fig:flow_budget} shows that micro-F1 increases as more flows are retained per visit trace. When only one flow is retained per visit trace, micro-F1 is 32.16\%, 37.31\%, and 36.24\% under Direct HTTPS, Trojan, and VMess, respectively. These results are consistent with website-identifying information being distributed across multiple flows and suggest that retaining more flows can improve window-level prediction.

\section{Conclusion}

We propose FlowAtom for multi-label WF in encrypted traffic with distinguishable connections. From a flow-level perspective, FlowAtom models each observation window as an unordered flow set and predicts the monitored website set without packet-level trace segmentation by visit or flow-to-website assignment at inference. FlowAtom learns flow representations from external unlabeled traffic and constructs shared Atoms from the training flows of each target traffic scenario. Max pooling of each Atom’s responses across flows yields a fixed-dimensional, permutation-invariant window representation for predicting the monitored website set. Across Direct HTTPS, Trojan, and VMess, FlowAtom achieves the highest micro-F1 among the evaluated methods in all 15 closed-world and 30 open-world settings tested.

\makeatletter
\let\oldthebibliography\thebibliography
\renewcommand{\thebibliography}[1]{%
  \oldthebibliography{#1}%
  \setlength{\itemsep}{0pt}%
  \setlength{\parskip}{0pt}%
}
\makeatother

\bibliographystyle{IEEEbib}
\bibliography{refs_com}

\end{document}